\PassOptionsToPackage{unicode=true}{hyperref} 
\PassOptionsToPackage{hyphens}{url}
\documentclass[]{article}
\usepackage{lmodern}
\usepackage{amssymb,amsmath}
\usepackage{ifxetex,ifluatex}
\usepackage{fixltx2e} 
\ifnum 0\ifxetex 1\fi\ifluatex 1\fi=0 
  \usepackage[T1]{fontenc}
  \usepackage[utf8]{inputenc}
  \usepackage{textcomp} 
\else 
  \usepackage{unicode-math}
  \defaultfontfeatures{Ligatures=TeX,Scale=MatchLowercase}
\fi
\IfFileExists{upquote.sty}{\usepackage{upquote}}{}
\IfFileExists{microtype.sty}{%
\usepackage[]{microtype}
\UseMicrotypeSet[protrusion]{basicmath} 
}{}
\IfFileExists{parskip.sty}{%
\usepackage{parskip}
}{
\setlength{\parindent}{0pt}
\setlength{\parskip}{6pt plus 2pt minus 1pt}
}
\usepackage{hyperref}
\hypersetup{
            pdftitle={INFERNO: Inference-Aware Neural Optimisation},
            pdfborder={0 0 0},
            breaklinks=true}
\usepackage{graphicx,grffile}
\makeatletter
\def\maxwidth{\ifdim\Gin@nat@width>\linewidth\linewidth\else\Gin@nat@width\fi}
\def\maxheight{\ifdim\Gin@nat@height>\textheight\textheight\else\Gin@nat@height\fi}
\makeatother
\setkeys{Gin}{width=\maxwidth,height=\maxheight,keepaspectratio}
\ifx\paragraph\undefined\else
\let\oldparagraph\paragraph
\renewcommand{\paragraph}[1]{\oldparagraph{#1}\mbox{}}
\fi
\ifx\subparagraph\undefined\else
\let\oldsubparagraph\subparagraph
\renewcommand{\subparagraph}[1]{\oldsubparagraph{#1}\mbox{}}
\fi

\makeatletter
\def\fps@figure{htbp}
\makeatother

\usepackage[nonatbib,preprint]{nips_2018}
\usepackage{booktabs}
\DeclareMathSymbol{\Gamma}{\mathord}{operators}{"00}
\DeclareMathSymbol{\Delta}{\mathord}{operators}{"01}
\DeclareMathSymbol{\Theta}{\mathord}{operators}{"02}
\DeclareMathSymbol{\Lambda}{\mathord}{operators}{"03}
\DeclareMathSymbol{\Xi}{\mathord}{operators}{"04}
\DeclareMathSymbol{\Pi}{\mathord}{operators}{"05}
\DeclareMathSymbol{\Sigma}{\mathord}{operators}{"06}
\DeclareMathSymbol{\Upsilon}{\mathord}{operators}{"07}
\DeclareMathSymbol{\Phi}{\mathord}{operators}{"08}
\DeclareMathSymbol{\Psi}{\mathord}{operators}{"09}
\DeclareMathSymbol{\Omega}{\mathord}{operators}{"0A}
\usepackage{algorithm}
\usepackage{algpseudocode}
\PassOptionsToPackage{sorting=none}{biblatex}
\makeatletter
\@ifpackageloaded{subfig}{}{\usepackage{subfig}}
\@ifpackageloaded{caption}{}{\usepackage{caption}}
\AtBeginDocument{%

}
\AtBeginDocument{%

}
\@ifpackageloaded{float}{}{\usepackage{float}}
\floatstyle{ruled}
\@ifundefined{c@chapter}{\newfloat{codelisting}{h}{lop}}{\newfloat{codelisting}{h}{lop}[chapter]}
\floatname{codelisting}{Listing}

\makeatother
\usepackage[]{biblatex}
\title{INFERNO: Inference-Aware Neural Optimisation}
\author{Pablo de Castro\\
INFN - Sezione di Padova\\
\texttt{pablo.de.castro@cern.ch} \And Tommaso Dorigo\\
INFN - Sezione di Padova\\
\texttt{tommaso.dorigo@cern.ch}}
\date{}

\begin{document}
\maketitle
\begin{abstract}
Complex computer simulations are commonly required for accurate data
modelling in many scientific disciplines, making statistical inference
challenging due to the intractability of the likelihood evaluation for
the observed data. Furthermore, sometimes one is interested on inference
drawn over a subset of the generative model parameters while taking into
account model uncertainty or misspecification on the remaining nuisance
parameters. In this work, we show how non-linear summary statistics can
be constructed by minimising inference-motivated losses via stochastic
gradient descent such they provided the smallest uncertainty for the
parameters of interest. As a use case, the problem of confidence
interval estimation for the mixture coefficient in a multi-dimensional
two-component mixture model (i.e.~signal vs background) is considered,
where the proposed technique clearly outperforms summary statistics
based on probabilistic classification, which are a commonly used
alternative but do not account for the presence of nuisance parameters.
\end{abstract}

\hypertarget{introduction}{%
\section{Introduction}\label{introduction}}

Simulator-based inference is currently at the core of many scientific
fields, such as population genetics, epidemiology, and experimental
particle physics. In many cases the implicit generative procedure
defined in the simulation is stochastic and/or lacks a tractable
probability density \(p(\boldsymbol{x}| \boldsymbol{\theta})\), where
\(\boldsymbol{\theta} \in \mathcal{\Theta}\) is the vector of model
parameters. Given some experimental observations
\(D = \{\boldsymbol{x}_0,...,\boldsymbol{x}_n\}\), a problem of special
relevance for these disciplines is statistical inference on a subset of
model parameters
\(\boldsymbol{\omega} \in \mathcal{\Omega} \subseteq \mathcal{\Theta}\).
This can be approached via likelihood-free inference algorithms such as
Approximate Bayesian Computation (ABC)
\autocite{beaumont2002approximate}, simplified synthetic likelihoods
\autocite{wood2010statistical} or density estimation-by-comparison
approaches \autocite{cranmer2015approximating}.

Because the relation between the parameters of the model and the data is
only available via forward simulation, most likelihood-free inference
algorithms tend to be computationally expensive due to the need of
repeated simulations to cover the parameter space. When data are
high-dimensional, likelihood-free inference can rapidly become
inefficient, so low-dimensional summary statistics \(\boldsymbol{s}(D)\)
are used instead of the raw data for tractability. The choice of summary
statistics for such cases becomes critical, given that naive choices
might cause loss of relevant information and a corresponding degradation
of the power of resulting statistical inference.

As a motivating example we consider data analyses at the Large Hadron
Collider (LHC), such as those carried out to establish the discovery of
the Higgs boson \autocites{higgs2012cms}{higgs2012atlas}. In that
framework, the ultimate aim is to extract information about Nature from
the large amounts of high-dimensional data on the subatomic particles
produced by energetic collision of protons, and acquired by highly
complex detectors built around the collision point. Accurate data
modelling is only available via stochastic simulation of a complicated
chain of physical processes, from the underlying fundamental interaction
to the subsequent particle interactions with the detector elements and
their readout. As a result, the density
\(p(\boldsymbol{x}| \boldsymbol{\theta})\) cannot be analytically
computed.

The inference problem in particle physics is commonly posed as
hypothesis testing based on the acquired data. An alternate hypothesis
\(H_1\) (e.g.~a new theory that predicts the existence of a new
fundamental particle) is tested against a null hypothesis \(H_0\)
(e.g.~an existing theory, which explains previous observed phenomena).
The aim is to check whether the null hypothesis can be rejected in
favour of the alternate hypothesis at a certain confidence level
surpassing \(1-\alpha\), where \(\alpha\), known as the Type I error
rate, is commonly set to \(\alpha=3\times10^{-7}\) for discovery claims.
Because \(\alpha\) is fixed, the sensitivity of an analysis is
determined by the power \(1-\beta\) of the test, where \(\beta\) is the
probability of rejecting a false null hypothesis, also known as Type II
error rate.

Due to the high dimensionality of the observed data, a low-dimensional
summary statistic has to be constructed in order to perform inference. A
well-known result of classical statistics, the Neyman-Pearson
lemma\autocite{NeymanPearson1933}, establishes that the likelihood-ratio
\(\Lambda(\boldsymbol{x})=p(\boldsymbol{x}| H_0)/p(\boldsymbol{x}| H_1)\)
is the most powerful test when two simple hypotheses are considered. As
\(p(\boldsymbol{x}| H_0)\) and \(p(\boldsymbol{x}| H_1)\) are not
available, simulated samples are used in practice to obtain an
approximation of the likelihood ratio by casting the problem as
supervised learning classification.

In many cases, the nature of the generative model (a mixture of
different processes) allows the treatment of the problem as signal (S)
vs background (B) classification \autocite{adam2015higgs}, when the task
becomes one of effectively estimating an approximation of
\(p_{S}(\boldsymbol{x})/p_{B}(\boldsymbol{x})\) which will vary
monotonically with the likelihood ratio. While the use of classifiers to
learn a summary statistic can be effective and increase the discovery
sensitivity, the simulations used to generate the samples which are
needed to train the classifier often depend on additional uncertain
parameters (commonly referred as nuisance parameters). These nuisance
parameters are not of immediate interest but have to be accounted for in
order to make quantitative statements about the model parameters based
on the available data. Classification-based summary statistics cannot
easily account for those effects, so their inference power is degraded
when nuisance parameters are finally taken into account.

In this work, we present a new machine learning method to construct
non-linear sample summary statistics that directly optimises the
expected amount of information about the subset of parameters of
interest using simulated samples, by explicitly and directly taking into
account the effect of nuisance parameters. In addition, the learned
summary statistics can be used to build synthetic sample-based
likelihoods and perform robust and efficient classical or Bayesian
inference from the observed data, so they can be readily applied in
place of current classification-based or domain-motivated summary
statistics in current scientific data analysis workflows.

\hypertarget{problem-statement}{%
\section{Problem Statement}\label{problem-statement}}

Let us consider a set of \(n\) i.i.d. observations
\(D = \{\boldsymbol{x}_0,...,\boldsymbol{x}_n\}\) where
\(\boldsymbol{x} \in \mathcal{X} \subseteq \mathbb{R}^d\), and a
generative model which implicitly defines a probability density
\(p(\boldsymbol{x} | \boldsymbol{\theta})\) used to model the data. The
generative model is a function of the vector of parameters
\(\boldsymbol{\theta} \in \mathcal{\Theta} \subseteq \mathbb{R}^p\),
which includes both relevant and nuisance parameters. We want to learn a
function
\(\boldsymbol{s} : \mathcal{D} \subseteq \mathbb{R}^{d\times n} \rightarrow \mathcal{S} \subseteq \mathbb{R}^{b}\)
that computes a summary statistic of the dataset and reduces its
dimensionality so likelihood-free inference methods can be applied
effectively. From here onwards, \(b\) will be used to denote the
dimensionality of the summary statistic \(\boldsymbol{s}(D)\).

While there might be infinite ways to construct a summary statistic
\(\boldsymbol{s} (D)\), we are only interested in those that are
informative about the subset of interest
\(\boldsymbol{\omega} \in \mathcal{\Omega} \subseteq \mathcal{\Theta}\)
of the model parameters. The concept of statistical sufficiency is
especially useful to evaluate whether summary statistics are
informative. In the absence of nuisance parameters, classical
sufficiency can be characterised by means of the factorisation
criterion: \begin{equation}
p(D|\boldsymbol{\omega}) = h(D) g(\boldsymbol{s}(D) | \boldsymbol{\omega} )
\label{eq:sufficiency}\end{equation} where \(h\) and \(g\) are
non-negative functions. If \(p(D | \boldsymbol{\omega})\) can be
factorised as indicated, the summary statistic \(\boldsymbol{s}(D)\)
will yield the same inference about the parameters
\(\boldsymbol{\omega}\) as the full set of observations \(D\). When
nuisance parameters have to be accounted in the inference procedure,
alternate notions of sufficiency are commonly used such as partial or
marginal sufficiency \autocites{basu2011partial}{sprott1975marginal}.
Nonetheless, for the problems of relevance in this work, the probability
density is not available in closed form so the general task of finding a
sufficient summary statistic cannot be tackled directly. Hence,
alternative methods to build summary statistics have to be followed.

For simplicity, let us consider a problem where we are only interested
on statistical inference on a single one-dimensional model parameter
\(\boldsymbol{\omega} = \{ \omega_0\}\) given some observed data. Be
given a summary statistic \(\boldsymbol{s}\) and a statistical procedure
to obtain an unbiased interval estimate of the parameter of interest
which accounts for the effect of nuisance parameters. The resulting
interval can be characterised by its width
\(\Delta \omega_0 = \hat{\omega}^{+}_0- \hat{\omega}^{-}_0\), defined by
some criterion so as to contain on average, upon repeated samping, a
given fraction of the probability density, e.g.~a central \(68.3\%\)
interval. The expected size of the interval depends on the summary
statistic \(\boldsymbol{s}\) chosen: in general, summary statistics that
are more informative about the parameters of interest will provide
narrower confidence or credible intervals on their value. Under this
figure of merit, the problem of choosing an optimal summary statistic
can be formally expressed as finding a summary statistic
\(\boldsymbol{s}^{\ast}\) that minimises the interval width:
\begin{equation}
\boldsymbol{s}^{\ast} = \textrm{argmin}_{\boldsymbol{s}}  \Delta \omega_0.
\label{eq:general_task}\end{equation} The above construction can be
extended to several parameters of interest by considering the interval
volume or any other function of the resulting confidence or credible
regions.

\hypertarget{sec:method}{%
\section{Method}\label{sec:method}}

\begin{figure}
\hypertarget{fig:diagram}{%
\centering
\includegraphics{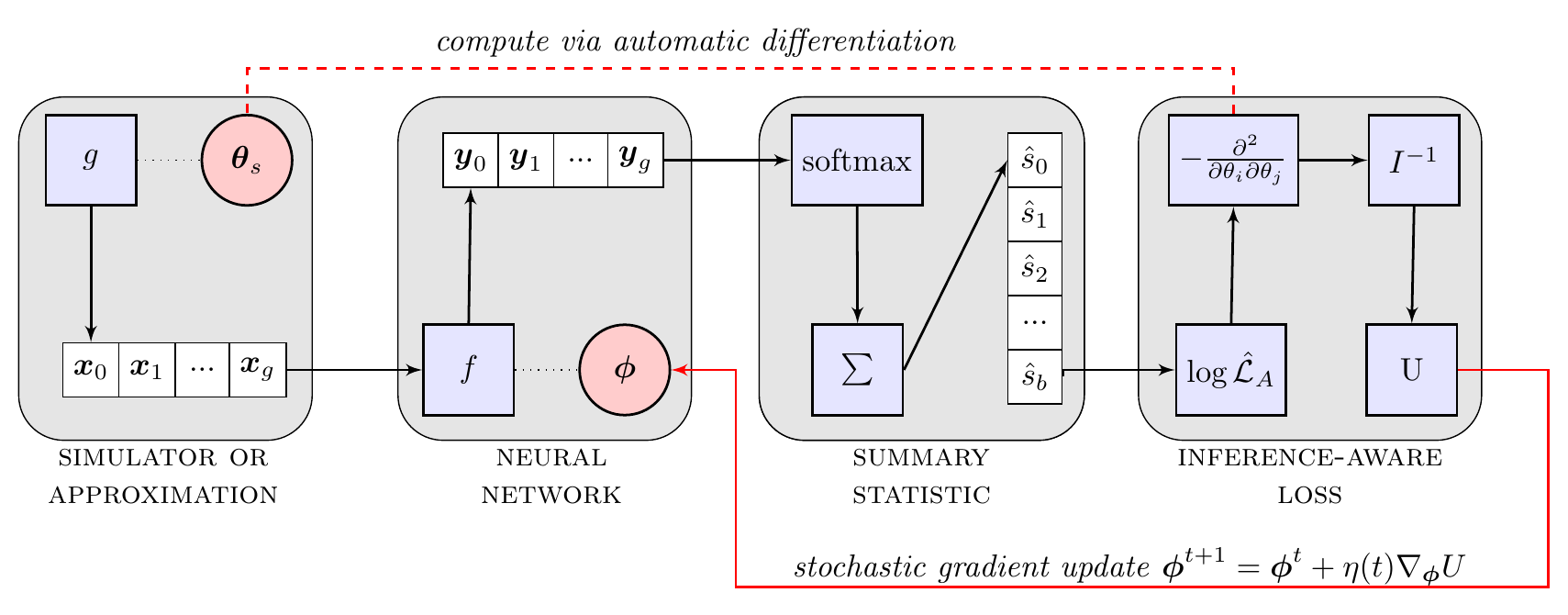}
\caption{Learning inference-aware summary statistics (see text for
details).}\label{fig:diagram}
}
\end{figure}

In this section, a machine learning technique to learn non-linear sample
summary statistics is described in detail. The method seeks to minimise
the expected variance of the parameters of interest obtained via a
non-parametric simulation-based synthetic likelihood. A graphical
description of the technique is depicted on Fig.~\ref{fig:diagram}. The
parameters of a neural network are optimised by stochastic gradient
descent within an automatic differentiation framework, where the
considered loss function accounts for the details of the statistical
model as well as the expected effect of nuisance parameters.

The family of summary statistics \(\boldsymbol{s}(D)\) considered in
this work is composed by a neural network model applied to each dataset
observation
\(\boldsymbol{f}(\boldsymbol{x}; \boldsymbol{\phi}) : \mathcal{X} \subseteq \mathbb{R}^{d} \rightarrow \mathcal{Y} \subseteq \mathbb{R}^{b}\),
whose parameters \(\boldsymbol{\phi}\) will be learned during training
by means of stochastic gradient descent, as will be discussed later.
Therefore, using set-builder notation the family of summary statistics
considered can be denoted as: \begin{equation}
\boldsymbol{s} (D, \boldsymbol{\phi})
 = \boldsymbol{s} \left ( \: \{ \:  \boldsymbol{f}(\boldsymbol{x}_i; \boldsymbol{\phi}) \:
  | \: \forall \: \boldsymbol{x}_i \in D \: \} \: \right )
\label{eq:summary}\end{equation} where
\(\boldsymbol{f}(\boldsymbol{x}_i; \boldsymbol{\phi})\) will reduce the
dimensionality from the input observations space \(\mathcal{X}\) to a
lower-dimensional space \(\mathcal{Y}\). The next step is to map
observation outputs to a dataset summary statistic, which will in turn
be calibrated and optimised via a non-parametric likelihood
\(\mathcal{L}(D; \boldsymbol{\theta},\boldsymbol{\phi})\) created using
a set of simulated observations
\(G_s= \{\boldsymbol{x}_0,...,\boldsymbol{x}_g\}\), generated at a
certain instantiation of the simulator parameters
\(\boldsymbol{\theta}_s\).

In experimental high energy physics experiments, which are the
scientific context that initially motivated this work, histograms of
observation counts are the most commonly used non-parametric density
estimator because the resulting likelihoods can be expressed as the
product of Poisson factors, one for each of the considered bins. A naive
sample summary statistic can be built from the output of the neural
network by simply assigning each observation \(\boldsymbol{x}\) to a bin
corresponding to the cardinality of the maximum element of
\(\boldsymbol{f}(\boldsymbol{x}; \boldsymbol{\phi})\), so each element
of the sample summary will correspond to the following sum:
\begin{equation}
s_i(D;\boldsymbol{\phi})=\sum_{\boldsymbol{x} \in D}
\begin{cases}
      1 & i = {argmax}_{j=\{0,...,b\}}
        (f_j(\boldsymbol{x}; \boldsymbol{\phi})) \\
      0 & i \neq {argmax}_{j=\{0,...,b\}}
        (f_j(\boldsymbol{x}; \boldsymbol{\phi})) \\
   \end{cases}
\label{eq:argmax}\end{equation} which can in turn be used to build the
following likelihood, where the expectation for each bin is taken from
the simulated sample \(G_s\): \begin{equation}
\mathcal{L}(D; \boldsymbol{\theta},\boldsymbol{\phi})=\prod_{i=0 }^b
             \textrm{Pois} \left ( s_i (D; \boldsymbol{\phi}) \:  |
             \: \left ( \frac{n}{g} \right ) s_i (G_s;\boldsymbol{\phi}) \right )
\label{eq:likelihood}\end{equation} where the \(n/g\) factor accounts
for the different number of observations in the simulated samples. In
cases where the number of observations is itself a random variable
providing information about the parameters of interest, or where the
simulated observation are weighted, the choice of normalisation of
\(\mathcal{L}\) may be slightly more involved and problem specific, but
nevertheless amenable.

In the above construction, the chosen family of summary statistics is
non-differentiable due to the \(argmax\) operator, so gradient-based
updates for the parameters cannot be computed. To work around this
problem, a differentiable approximation
\(\hat{\boldsymbol{s}}(D ; \boldsymbol{\phi})\) is considered. This
function is defined by means of a \(softmax\) operator: \begin{equation}
\hat{s}_i(D;\boldsymbol{\phi})=\sum_{x \in D}
  \frac{e^{f_i(\boldsymbol{x}; \boldsymbol{\phi})/\tau}}
  {\sum_{j=0}^{b} e^{f_j(\boldsymbol{x}; \boldsymbol{\phi})/\tau}}
\label{eq:soft_summary}\end{equation} where the temperature
hyper-parameter \(\tau\) will regulate the softness of the operator. In
the limit of \(\tau \rightarrow 0^{+}\), the probability of the largest
component will tend to 1 while others to 0, and therefore
\(\hat{\boldsymbol{s}}(D ; \boldsymbol{\phi}) \rightarrow \boldsymbol{s}(D; \boldsymbol{\phi})\).
Similarly, let us denote by
\(\hat{\mathcal{L}}(D; \boldsymbol{\theta}, \boldsymbol{\phi})\) the
differentiable approximation of the non-parametric likelihood obtained
by substituting \(\boldsymbol{s}(D ; \boldsymbol{\phi})\) with
\(\hat{\boldsymbol{s}}(D ; \boldsymbol{\phi})\). Instead of using the
observed data \(D\), the value of \(\hat{\mathcal{L}}\) may be computed
when the observation for each bin is equal to its corresponding
expectation based on the simulated sample \(G_s\), which is commonly
denoted as the Asimov likelihood \autocite{cowan2011asymptotic}
\(\hat{\mathcal{L}}_A\): \begin{equation}
\hat{\mathcal{L}}_A(\boldsymbol{\theta}; \boldsymbol{\phi})=\prod_{i=0 }^b
             \textrm{Pois} \left ( \left ( \frac{n}{g} \right )
            \hat{s}_i (G_s;\boldsymbol{\phi}) \:  | \: \left ( \frac{n}{g} \right )
             \hat{s}_i (G_s;\boldsymbol{\phi}) \right )
\label{eq:likelihood_asimov}\end{equation} for which it can be easily
proven that
\(argmax_{\boldsymbol{\theta} \in \mathcal{\theta}} (\hat{\mathcal{L}}_A( \boldsymbol{\theta; \boldsymbol{\phi}})) = \boldsymbol{\theta}_s\),
so the maximum likelihood estimator (MLE) for the Asimov likelihood is
the parameter vector \(\boldsymbol{\theta}_s\) used to generate the
simulated dataset \(G_s\). In Bayesian terms, if the prior over the
parameters is flat in the chosen metric, then \(\boldsymbol{\theta}_s\)
is also the maximum a posteriori (MAP) estimator. By taking the negative
logarithm and expanding in \(\boldsymbol{\theta}\) around
\(\boldsymbol{\theta}_s\), we can obtain the Fisher information matrix
\autocite{fisher_1925} for the Asimov likelihood: \begin{equation}
{\boldsymbol{I}(\boldsymbol{\theta})}_{ij}
= \frac{\partial^2}{\partial {\theta_i} \partial {\theta_j}}
 \left ( - \log \mathcal{\hat{L}}_A(\boldsymbol{\theta};
 \boldsymbol{\phi}) \right )
\label{eq:fisher_info}\end{equation} which can be computed via automatic
differentiation if the simulation is differentiable and included in the
computation graph or if the effect of varying \(\boldsymbol{\theta}\)
over the simulated dataset \(G_s\) can be effectively approximated.
While this requirement does constrain the applicability of the proposed
technique to a subset of likelihood-free inference problems, it is quite
common for e.g.~physical sciences that the effect of the parameters of
interest and the main nuisance parameters over a sample can be
approximated by the changes of mixture coefficients of mixture models,
translations of a subset of features, or conditional density ratio
re-weighting.

If \(\hat{\boldsymbol{\theta}}\) is an unbiased estimator of the values
of \(\boldsymbol{\theta}\), the covariance matrix fulfils the Cramér-Rao
lower bound \autocites{cramer2016mathematical}{rao1992information}:
\begin{equation}
\textrm{cov}_{\boldsymbol{\theta}}(\hat{\boldsymbol{\theta}}) \geq
I(\boldsymbol{\theta})^{-1}
\label{eq:CRB}\end{equation} and the inverse of the Fisher information
can be used as an approximate estimator of the expected variance, given
that the bound would become an equality in the asymptotic limit for MLE.
If some of the parameters \(\boldsymbol{\theta}\) are constrained by
independent measurements characterised by their likelihoods
\(\{\mathcal{L}_C^{0}(\boldsymbol{\theta}), ..., \mathcal{L}_{C}^{c}(\boldsymbol{\theta})\}\),
those constraints can also be easily included in the covariance
estimation, simply by considering the augmented likelihood
\(\hat{\mathcal{L}}_A'\) instead of \(\hat{\mathcal{L}}_A\) in
Eq.~\ref{eq:fisher_info}:
\begin{equation}\hat{\mathcal{L}}_A'(\boldsymbol{\theta} ; \boldsymbol{\phi}) =
\hat{\mathcal{L}}_A(\boldsymbol{\theta} ; \boldsymbol{\phi})
\prod_{i=0}^{c}\mathcal{L}_C^i(\boldsymbol{\theta}).\label{eq:add_constraint}\end{equation}
In Bayesian terminology, this approach is referred to as the Laplace
approximation \autocite{laplace1986memoir} where the logarithm of the
joint density (including the priors) is expanded around the MAP to a
multi-dimensional normal approximation of the posterior density:
\begin{equation}
p(\boldsymbol{\theta}|D) \approx \textrm{Normal}(
\boldsymbol{\theta} ; \hat{\boldsymbol{\theta}},
I(\hat{\boldsymbol{\theta})}^{-1} )
\label{eq:normal_approx}\end{equation} which has already been approached
by automatic differentiation in probabilistic programming frameworks
\autocite{tran2016edward}. While a histogram has been used to construct
a Poisson count sample likelihood, non-parametric density estimation
techniques can be used in its place to construct a product of
observation likelihoods based on the neural network output
\(\boldsymbol{f}(\boldsymbol{x}; \boldsymbol{\phi})\) instead. For
example, an extension of this technique to use kernel density estimation
(KDE) should be straightforward, given its intrinsic differentiability.

The loss function used for stochastic optimisation of the neural network
parameters \(\boldsymbol{\phi}\) can be any function of the inverse of
the Fisher information matrix at \(\boldsymbol{\theta}_s\), depending on
the ultimate inference aim. The diagonal elements
\(I_{ii}^{-1}(\boldsymbol{\theta}_s)\) correspond to the expected
variance of each of the \(\phi_i\) under the normal approximation
mentioned before, so if the aim is efficient inference about one of the
parameters \(\omega_0 = \theta_k\) a candidate loss function is:
\begin{equation}
U = I_{kk}^{-1}(\boldsymbol{\theta}_s)
\label{eq:example_loss}\end{equation} which corresponds to the expected
width of the confidence interval for \(\omega_0\) accounting also for
the effect of the other nuisance parameters in \(\boldsymbol{\theta}\).
This approach can also be extended when the goal is inference over
several parameters of interest
\(\boldsymbol{\omega} \subseteq \boldsymbol{\theta}\) (e.g.~when
considering a weighted sum of the relevant variances). A simple version
of the approach just described to learn a neural-network based summary
statistic employing an inference-aware loss is summarised in Algorithm
\autoref{alg:simple_algorithm}.

\begin{algorithm}[H]
  \caption{Inference-Aware Neural Optimisation.}
  \begin{flushleft}
    {\it Input 1:} differentiable simulator or variational
    approximation $g(\boldsymbol{\theta})$. \\
    {\it Input 2:} initial parameter values $\boldsymbol{\theta}_s.$ \\
    {\it Input 3:} parameter of interest $\omega_0=\theta_k$.
     \\
    {\it Output:} learned summary statistic
      $\boldsymbol{s}(D; \boldsymbol{\phi})$.\\
 \end{flushleft}
 \begin{algorithmic}[1]
 \For{$i=1$ to $N$}
  \State{Sample a representative mini-batch $G_s$ from
  $g(\boldsymbol{\theta}_s)$.}
  \State{Compute differentiable summary statistic
    $\hat{\boldsymbol{s}}(G_s;\boldsymbol{\phi})$.}
  \State{Construct Asimov likelihood
    $\mathcal{L}_A(\boldsymbol{\theta}, \boldsymbol{\phi})$.}
  \State{Get information matrix inverse $I(\boldsymbol{\theta})^{-1}
  = \boldsymbol{H}_{\boldsymbol{\theta}}^{-1}(\log
  \mathcal{L}_A(\boldsymbol{\theta}, \boldsymbol{\phi}))$.}
  \State{Obtain loss
    $U= I_{kk}^{-1}(\boldsymbol{\theta}_s)$.}
  \State{Update network parameters $\boldsymbol{\phi} \rightarrow
  \textrm{SGD}(\nabla_{\boldsymbol{\phi}} U)$.}  
 \EndFor
 \end{algorithmic}
 \label{alg:simple_algorithm}
\end{algorithm}

\hypertarget{related-work}{%
\section{Related Work}\label{related-work}}

Classification or regression models have been implicitly used to
construct summary statistics for inference in several scientific
disciplines. For example, in experimental particle physics, the mixture
model structure of the problem makes it amenable to supervised
classification based on simulated datasets
\autocites{hocker2007tmva}{baldi2014searching}. While a classification
objective can be used to learn powerful feature representations and
increase the sensitivity of an analysis, it does not take into account
the details of the inference procedure or the effect of nuisance
parameters like the solution proposed in this work.

The first known effort to include the effect of nuisance parameters in
classification and explain the relation between classification and the
likelihood ratio was by Neal \autocite{neal2008computing}. In the
mentioned work, Neal proposes training of classifier including a
function of nuisance parameter as additional input together with a
per-observation regression model of the expectation value for inference.
Cranmer et al. \autocite{cranmer2015approximating} improved on this
concept by using a parametrised classifier to approximate the likelihood
ratio which is then calibrated to perform statistical inference. At
variance with the mentioned works, we do not consider a classification
objective at all and the neural network is directly optimised based on
an inference-aware loss. Additionally, once the summary statistic has
been learnt the likelihood can be trivially constructed and used for
classical or Bayesian inference without a dedicated calibration step.
Furthermore, the approach presented in this work can also be extended,
as done by Baldi et al. \autocite{baldi2016parameterized} by a subset of
the inference parameters to obtain a parametrised family of summary
statistics with a single model.

Recently, Brehmer et al.
\autocites{Brehmer:2018hga}{brehmer2018constraining}{brehmer2018guide}
further extended the approach of parametrised classifiers to better
exploit the latent-space space structure of generative models from
complex scientific simulators. Additionally they propose a family of
approaches that include a direct regression of the likelihood ratio
and/or likelihood score in the training losses. While extremely
promising, the most performing solutions are designed for a subset of
the inference problems at the LHC and they require considerable changes
in the way the inference is carried out. The aim of this work is
different, as we try to learn sample summary statistics that may act as
a plug-in replacement of classifier-based dimensionality reduction and
can be applied to general likelihood-free problems where the effect of
the parameters can be modelled or approximated.

Within the field of Approximate Bayesian Computation (ABC), there have
been some attempts to use neural network as a dimensionality reduction
step to generate summary statistics. For example, Jiang et al.
\autocite{jiang2015learning} successfully employ a summary statistic by
directly regressing the parameters of interest and therefore
approximating the posterior mean given the data, which then can be used
directly as a summary statistic.

A different path is taken by Louppe et al.
\autocite{louppe2017learning}, where the authors present a adversarial
training procedure to enforce a pivotal property on a predictive model.
The main concern of this approach is that a classifier which is pivotal
with respect to nuisance parameters might not be optimal, neither for
classification nor for statistical inference. Instead of aiming for
being pivotal, the summary statistics learnt by our algorithm attempt to
find a transformation that directly reduces the expected effect of
nuisance parameters over the parameters of interest.

\hypertarget{experiments}{%
\section{Experiments}\label{experiments}}

In this section, we first study the effectiveness of the inference-aware
optimisation in a synthetic mixture problem where the likelihood is
known. We then compare our results with those obtained by standard
classification-based summary statistics. All the code needed to
reproduce the results presented the results presented here is available
in an online repository \autocite{code_repository}, extensively using
\textsc{TensorFlow} \autocite{tensorflow2015-whitepaper} and
\textsc{TensorFlow Probability}
\autocites{tran2016edward}{dillon2017tensorflow} software libraries.

\hypertarget{d-synthetic-mixture}{%
\subsection{3D Synthetic Mixture}\label{d-synthetic-mixture}}

\begin{figure}
\centering

\subfloat[signal (red) and background
(blue)]{\includegraphics[width=0.49\textwidth,height=\textheight]{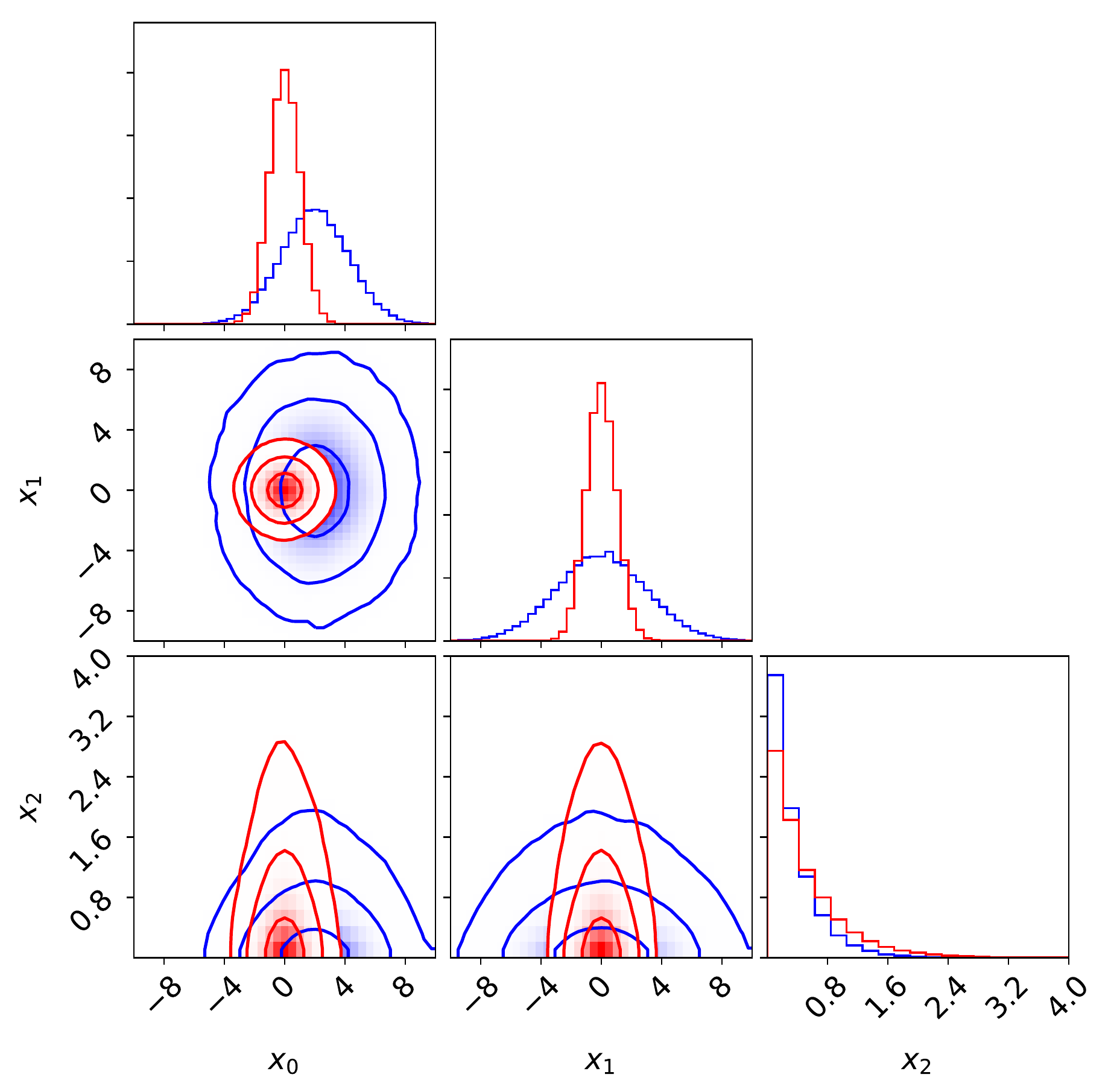}\label{fig:subfigure_a}}
\subfloat[mixture distribution
(black)]{\includegraphics[width=0.49\textwidth,height=\textheight]{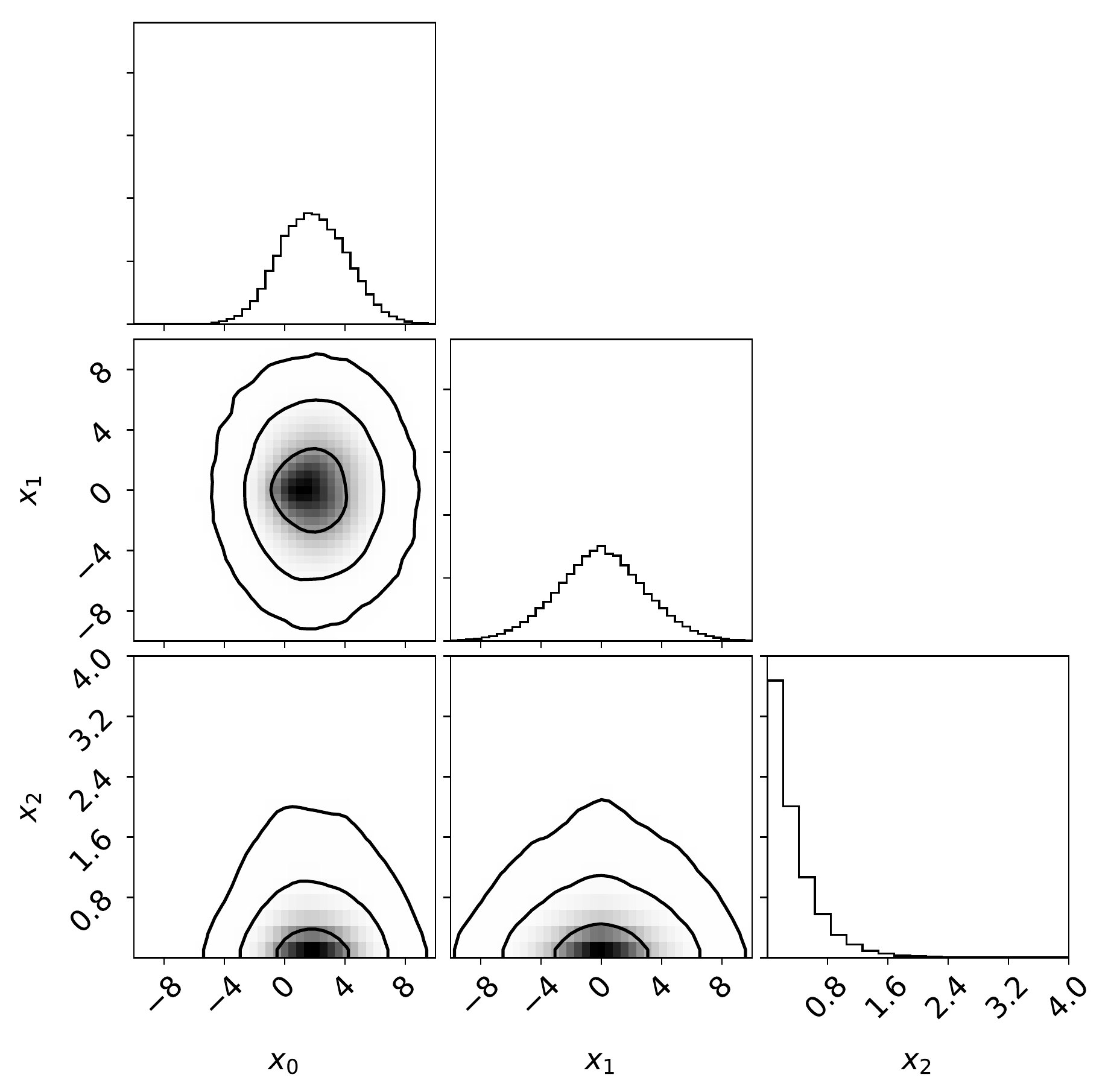}\label{fig:subfigure_b}}

\caption{Projection in 1D and 2D dimensions of 50000 samples from the
synthetic problem considered. The background distribution nuisance
parameters used for generating data correspond to \(r=0\) and
\(\lambda=3\). For samples the mixture distribution, \(s=50\) and
\(b=1000\) were used, hence the mixture coefficient is \(\mu=50/1050\).}

\label{fig:subfigs_distributions}

\end{figure}

In order to exemplify the usage of the proposed approach, evaluate its
viability and test its performance by comparing to the use of a
classification model proxy, a three-dimensional mixture example with two
components is considered. One component will be referred as background
\(f_b(\boldsymbol{x} | \lambda)\) and the other as signal
\(f_s(\boldsymbol{x})\); their probability density functions are taken
to correspond respectively to: \begin{equation}
f_b(\boldsymbol{x} | r, \lambda) =
\mathcal{N} \left (
  (x_0, x_1) \, \middle | \,
  (2+r, 0),
  \begin{bmatrix}
    5 & 0 \\
    0 & 9 \\
   \end{bmatrix}
\right)
Exp (x_2 | \lambda)
\label{eq:bkg_toy_pdf}\end{equation} \begin{equation}
f_s(\boldsymbol{x}) =
\mathcal{N} \left (
  (x_0, x_1) \, \middle | \,
  (1,1),
  \begin{bmatrix}
    1 & 0 \\
    0 & 1 \\
   \end{bmatrix}
\right)
Exp (x_2 | 2)
\label{eq:sig_toy_pdf}\end{equation} so that \((x_0,x_1)\) are
distributed according to a multivariate normal distribution while
\(x_2\) follows an independent exponential distribution both for
background and signal, as shown in Fig.~\ref{fig:subfigure_a}. The
signal distribution is fully specified while the background distribution
depends on \(r\), a parameter which shifts the mean of the background
density, and a parameter \(\lambda\) which specifies the exponential
rate in the third dimension. These parameters will be the treated as
nuisance parameters when benchmarking different methods. Hence, the
probability density function of observations has the following mixture
structure: \begin{equation}
p(\boldsymbol{x}| \mu, r, \lambda) = (1-\mu) f_b(\boldsymbol{x} | r, \lambda) 
                                      + \mu f_s(\boldsymbol{x})
\label{eq:mixture_eq}\end{equation} where \(\mu\) is the parameter
corresponding to the mixture weight for the signal and consequently
\((1-\mu)\) is the mixture weight for the background. The
low-dimensional projections from samples from the mixture distribution
for a small \(\mu=50/1050\) is shown in Fig.~\ref{fig:subfigure_b}.

Let us assume that we want to carry out inference based on \(n\) i.i.d.
observations, such that \(\mathbb{E}[n_s]=\mu n\) observations of signal
and \(\mathbb{E}[n_b] = (1-\mu)n\) observations of background are
expected, respectively. While the mixture model parametrisation shown in
Eq.~\ref{eq:mixture_eq} is correct as is, the underlying model could
also give information on the expected number of observations as a
function of the model parameters. In this toy problem, we consider a
case where the underlying model predicts that the total number of
observations are Poisson distributed with a mean \(s+b\), where \(s\)
and \(b\) are the expected number of signal and background observations.
Thus the following parametrisation will be more convenient for building
sample-based likelihoods: \begin{equation}
p(\boldsymbol{x}| s, r, \lambda, b) = \frac{b}{ s+b}
 f_b(\boldsymbol{x} | r, \lambda) +
 \frac{s}{s+b} f_s(\boldsymbol{x}).
\label{eq:mixture_alt}\end{equation} This parametrisation is common for
physics analyses at the LHC, because theoretical calculations provide
information about the expected number of observations. If the
probability density is known, but the expectation for the number of
observed events depends on the model parameters, the likelihood can be
extended \autocite{barlow1990extended} with a Poisson count term as:
\begin{equation}
\mathcal{L}(s, r, \lambda, b) = \textrm{Pois}(n | s+b) \prod^{n}
p(\boldsymbol{x}| s,r, \lambda, b)
\label{eq:ext_ll}\end{equation} which will be used to provide an optimal
inference baseline when benchmarking the different approaches. Another
quantity of relevance is the conditional density ratio, which would
correspond to the optimal classifier (in the Bayes risk sense)
separating signal and background events in a balanced dataset (equal
priors): \begin{equation}
s^{*}(\boldsymbol{x} | r, \lambda) = \frac{f_s(\boldsymbol{x})}{f_s(\boldsymbol{x}) + f_b(\boldsymbol{x} | r, \lambda) }
\label{eq:opt_clf}\end{equation} noting that this quantity depends on
the parameters that define the background distribution \(r\) and
\(\lambda\), but not on \(s\) or \(b\) that are a function of the
mixture coefficients. It can be proven (see appendix
\ref{sec:sufficiency} ) that \(s^{*}(\boldsymbol{x})\) is a sufficient
summary statistic with respect to an arbitrary two-component mixture
model if the only unknown parameter is the signal mixture fraction
\(\mu\) (or alternatively \(s\) in the chosen parametrisation). In
practise, the probability density functions of signal and background are
not known analytically, and only forward samples are available through
simulation, so alternative approaches are required.

While the synthetic nature of this example allows to rapidly generate
training data on demand, a training dataset of 200,000 simulated
observations has been considered, in order to study how the proposed
method performs when training data is limited. Half of the simulated
observations correspond to the signal component and half to the
background component. The latter has been generated using \(r=0.0\) and
\(\lambda=3.0\). A validation holdout from the training dataset of
200,000 observations is only used for computing relevant metrics during
training and to control over-fitting. The final figures of merit that
allow to compare different approaches are computed using a larger
dataset of 1,000,000 observations. For simplicity, mini-batches for each
training step are balanced so the same number of events from each
component is taken both when using standard classification or
inference-aware losses.

An option is to pose the problem as one of classification based on a
simulated dataset. A supervised machine learning model such a neural
network can be trained to discriminate signal and background
observations, considering a fixed parameters \(r\) and \(\lambda\). The
output of such a model typically consist in class probabilities \(c_s\)
and \(c_b\) given an observation \(\boldsymbol{x}\), which will tend
asymptotically to the optimal classifier from Eq.~\ref{eq:opt_clf} given
enough data, a flexible enough model and a powerful learning rule. The
conditional class probabilities (or alternatively the likelihood ratio
\(f_s(\boldsymbol{x})/f_b(\boldsymbol{x})\)) are powerful learned
features that can be used as summary statistic; however their
construction ignores the effect of the nuisance parameters \(r\) and
\(\lambda\) on the background distribution. Furthermore, some kind of
non-parametric density estimation (e.g.~a histogram) has to be
considered in order to build a calibrated statistical model using the
classification-based learned features, which will in turn smooth and
reduce the information available for inference.

\begin{figure}
\centering

\subfloat[classifier trained on simulated
samples]{\includegraphics[width=0.48\textwidth,height=\textheight]{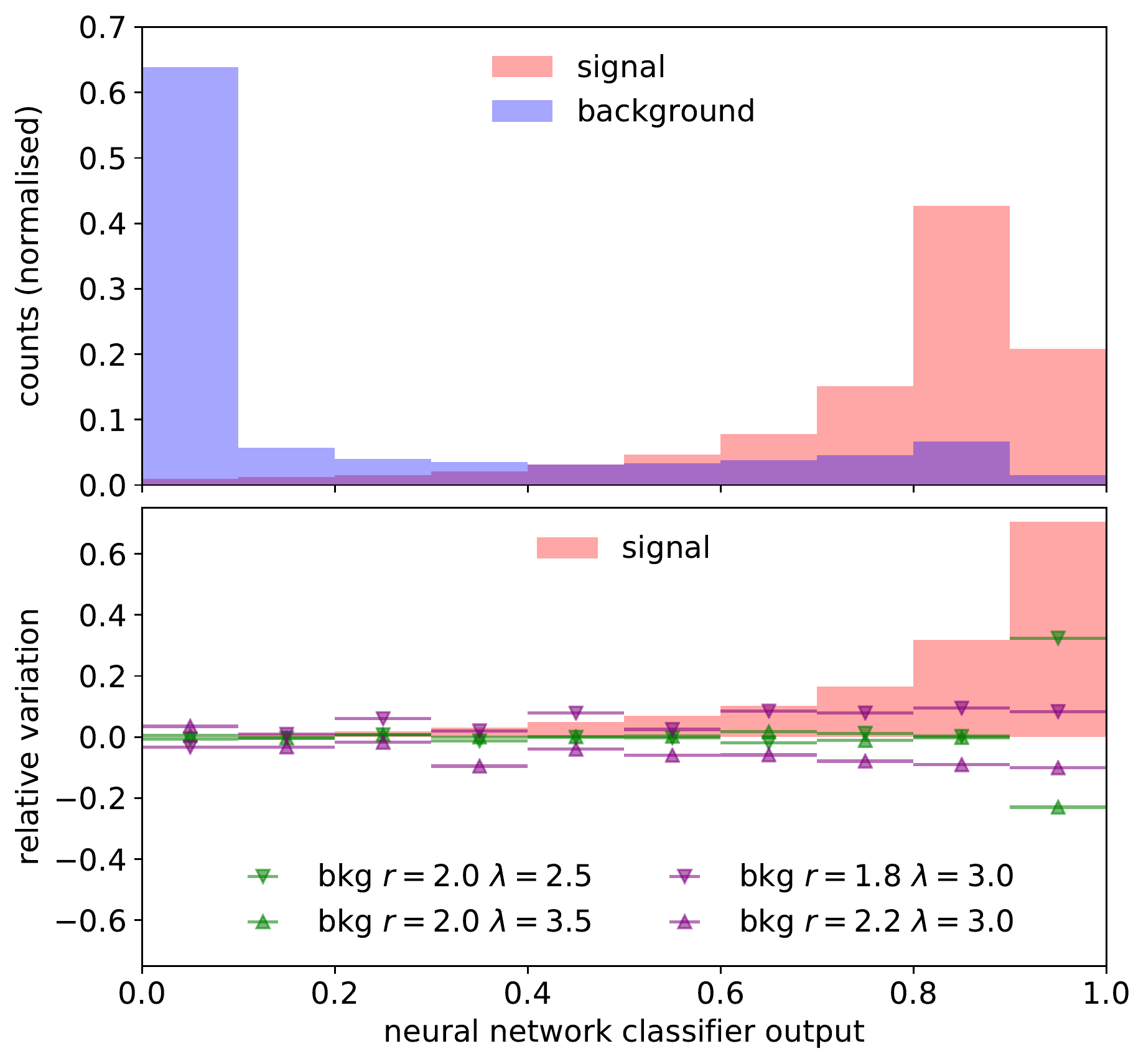}\label{fig:train_clf}}
\subfloat[optimal classifier
\(s(\boldsymbol{x} | r = 0.0, \lambda = 3.0)\)]{\includegraphics[width=0.48\textwidth,height=\textheight]{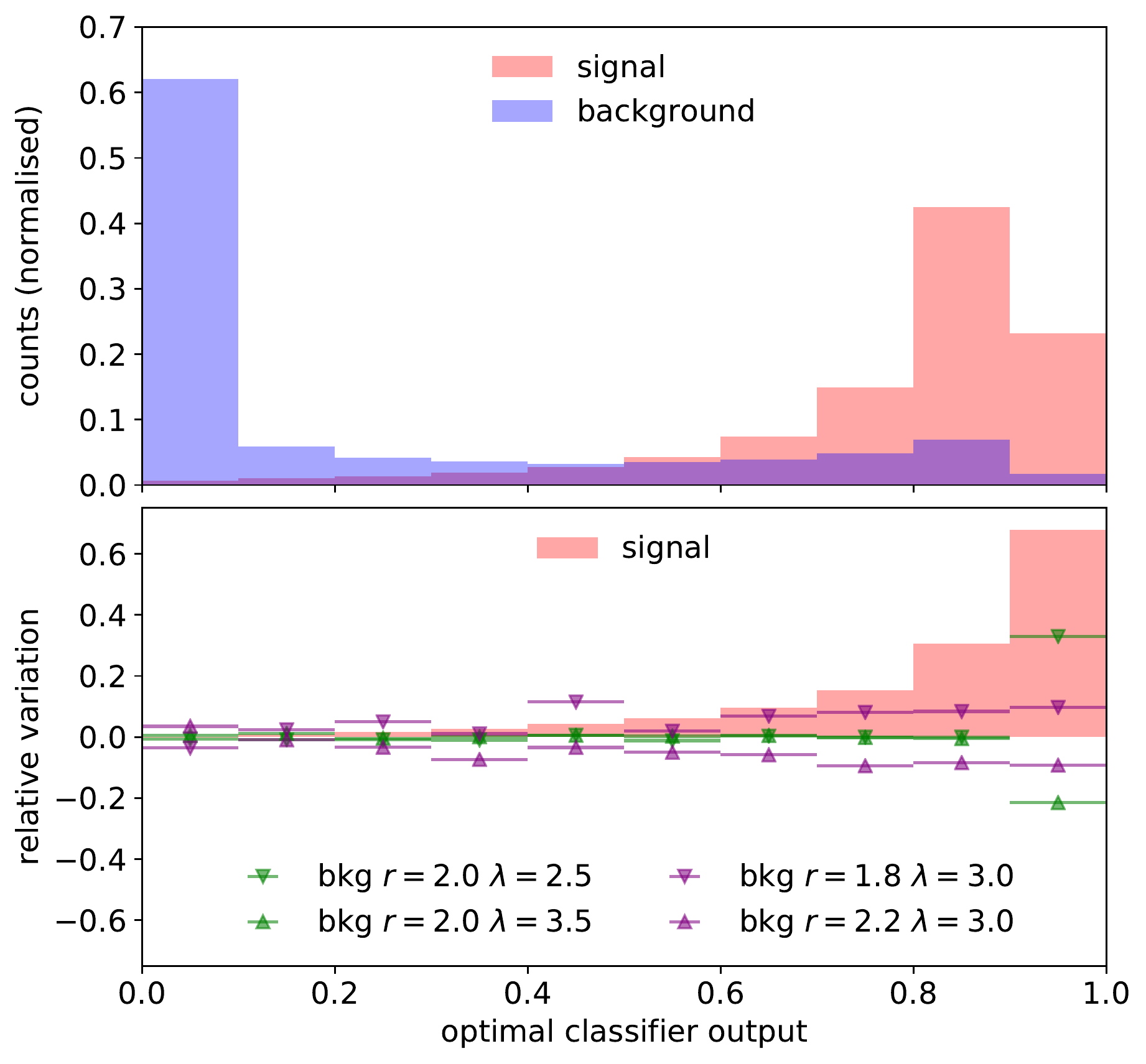}\label{fig:opt_clf}}

\caption{Histograms of summary statistics for signal and background
(top) and variation for different values of nuisance parameters compared
with the expected signal relative to the nominal background magniture
(bottom). The classifier was trained using signal and background samples
generated for \(r = 0.0\) and \(\lambda = 3.0\).}

\label{fig:subfigs_clf_hists}

\end{figure}

To exemplify the use of this family of classification-based summary
statistics, a histogram of a deep neural network classifier output
trained on simulated data and its variation computed for different
values of \(r\) and \(\lambda\) are shown in Fig.~\ref{fig:train_clf}.
The details of the training procedure will be provided later in this
document. The classifier output can be directly compared with
\(s(\boldsymbol{x} | r = 0.0, \lambda = 3.0)\) evaluated using the
analytical distribution function of signal and background according to
Eq.~\ref{eq:opt_clf}, which is shown in Fig.~\ref{fig:opt_clf} and
corresponds to the optimal classifier. The trained classifier
approximates very well the optimal classifier. The summary statistic
distribution for background depends considerably on the value of the
nuisance parameters both for the trained and the optimal classifier,
which will in turn cause an important degradation on the subsequent
statistical inference.

The statistical model described above has up to four unknown parameters:
the expected number of signal observations \(s\), the background mean
shift \(r\), the background exponential rate in the third dimension
\(\lambda\), and the expected number of background observations. The
effect of the expected number of signal and background observations
\(s\) and \(b\) can be easily included in the computation graph by
weighting the signal and background observations. This is equivalent to
scaling the resulting vector of Poisson counts (or its differentiable
approximation) if a non-parametric counting model as the one described
in Sec.~\ref{sec:method} is used. Instead the effect of \(r\) and
\(\lambda\), both nuisance parameters that will define the background
distribution, is more easily modelled as a transformation of the input
data \(\boldsymbol{x}\). In particular, \(r\) is a nuisance parameter
that causes a shift on the background along the first dimension and its
effect can accounted for in the computation graph by simply adding
\((r,0.0,0.0)\) to each observation in the mini-batch generated from the
background distribution. Similarly, the effect of \(\lambda\) can be
modelled by multiplying \(x_2\) by the ratio between the \(\lambda_0\)
used for generation and the one being modelled. These transformations
are specific for this example, but alternative transformations depending
on parameters could also be accounted for as long as they are
differentiable or substituted by a differentiable approximation.

For this problem, we are interested in carrying out statistical
inference on the parameter of interest \(s\). In fact, the performance
of inference-aware optimisation as described in Sec.~\ref{sec:method}
will be compared with classification-based summary statistics for a
series of inference benchmarks based on the synthetic problem described
above that vary in the number of nuisance parameters considered and
their constraints:

\begin{itemize}
  \item \textbf{Benchmark 0:} no nuisance parameters are considered, both signal and
  background distributions are taken as fully specified ($r=0.0$, $\lambda=3.0$
  and $b=1000.$).
  \item \textbf{Benchmark 1:} $r$ is considered as an unconstrained
  nuisance parameter, while $\lambda=3.0$ and $b=1000$ are fixed. 
  \item \textbf{Benchmark 2:} $r$ and $\lambda$ are considered as unconstrained
  nuisance parameters, while $b=1000$ is fixed. 
  \item \textbf{Benchmark 3:} $r$ and $\lambda$ are considered as
  nuisance parameters but with the following constraints: $\mathcal{N} (r |0.0, 0.4)$
  and $\mathcal{N} (\lambda| 3.0, 1.0)$, while $b=1000$ is fixed. 
  \item \textbf{Benchmark 4:} all $r$, $\lambda$ and $b$ are all considered as
  nuisance parameters with the following constraints: $\mathcal{N} (r |0.0, 0.4)$,
  $\mathcal{N} (\lambda| 3.0, 1.0)$ and $\mathcal{N} (b | 1000., 100.)$ . 
\end{itemize}

When using classification-based summary statistics, the construction of
a summary statistic does depend on the presence of nuisance parameters,
so the same model is trained independently of the benchmark considered.
In real-world inference scenarios, nuisance parameters have often to be
accounted for and typically are constrained by prior information or
auxiliary measurements. For the approach presented in this work,
inference-aware neural optimisation, the effect of the nuisance
parameters and their constraints can be taken into account during
training. Hence, 5 different training procedures for \textsc{INFERNO}
will be considered, one for each of the benchmarks, denoted by the same
number.

The same basic network architecture is used both for cross-entropy and
inference-aware training: two hidden layers of 100 nodes followed by
ReLU activations. The number of nodes on the output layer is two when
classification proxies are used, matching the number of mixture classes
in the problem considered. Instead, for inference-aware classification
the number of output nodes can be arbitrary and will be denoted with
\(b\), corresponding to the dimensionality of the sample summary
statistics. The final layer is followed by a softmax activation function
and a temperature \(\tau = 0.1\) for inference-aware learning to ensure
that the differentiable approximations are closer to the true
expectations. Standard mini-batch stochastic gradient descent (SGD) is
used for training and the optimal learning rate is fixed and decided by
means of a simple scan; the best choice found is specified together with
the results.

\begin{figure}
\centering

\subfloat[inference-aware training
loss]{\includegraphics[width=0.48\textwidth,height=\textheight]{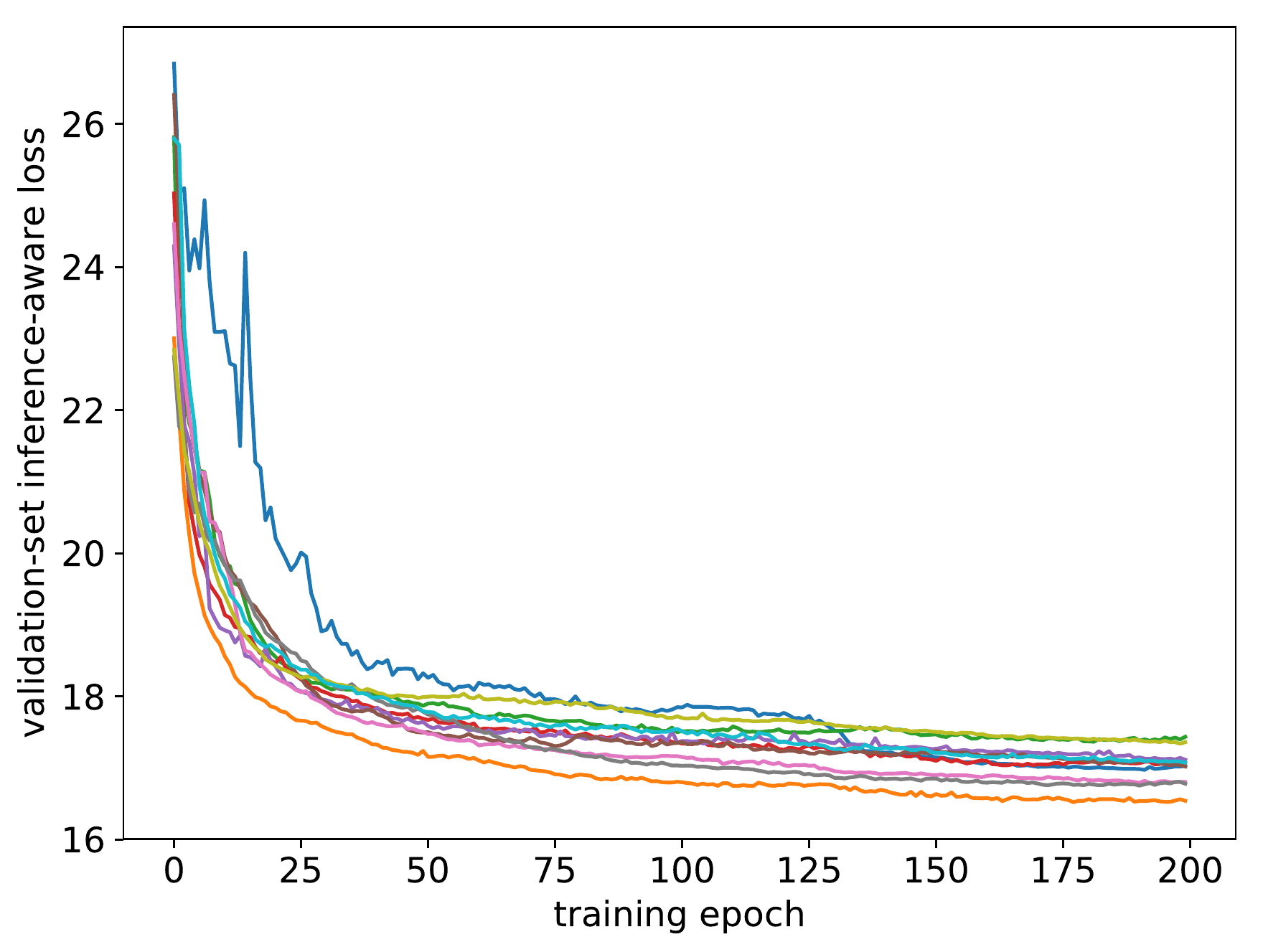}\label{fig:training_dynamics}}
\subfloat[profile-likelihood
comparison]{\includegraphics[width=0.48\textwidth,height=\textheight]{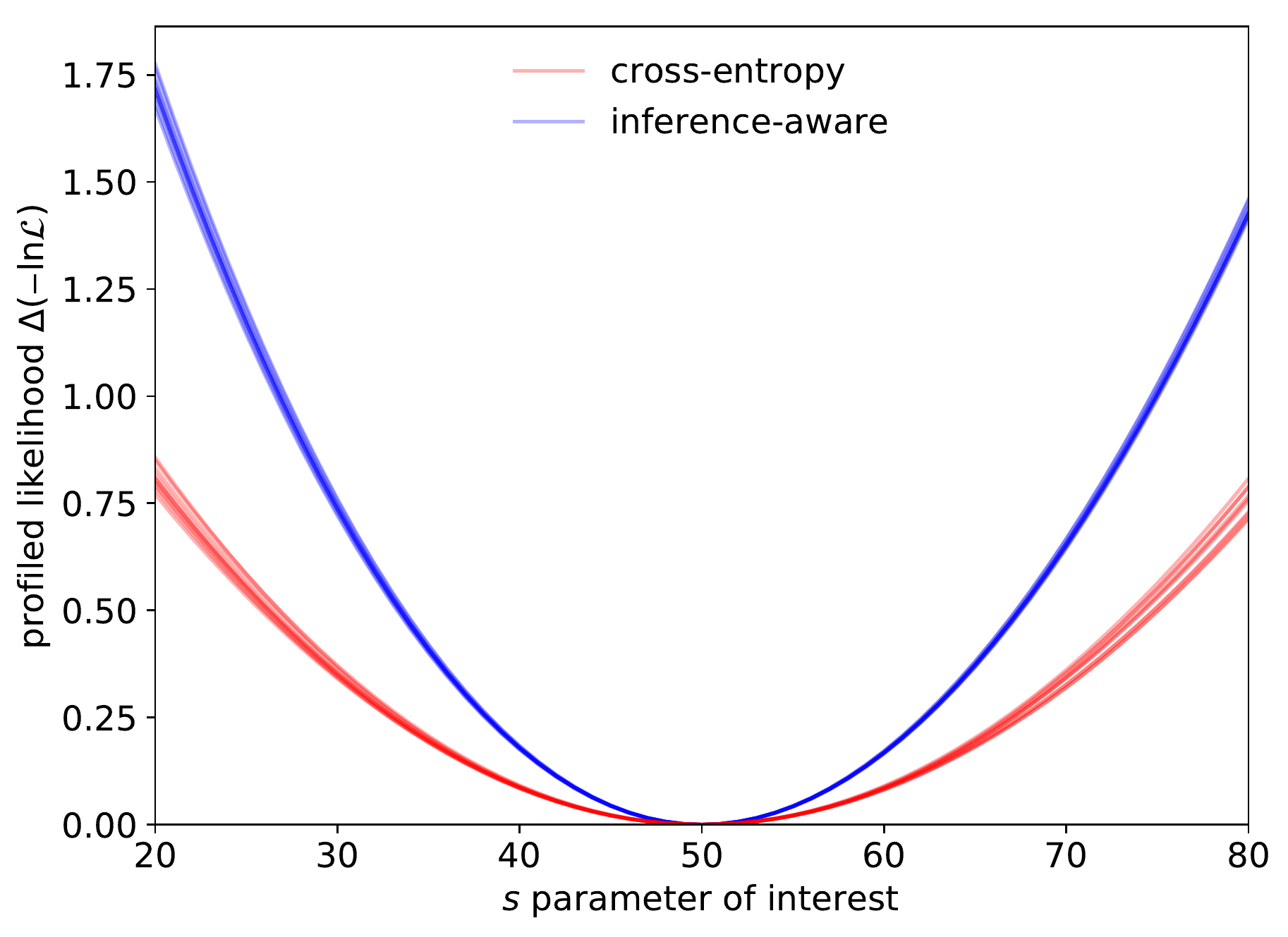}\label{fig:profile_likelihood}}

\caption{Dynamics and results of inference-aware optimisation: (a)
square root of inference-loss (i.e.~approximated standard deviation of
the parameter of interest) as a function of the training step for 10
different random initialisations of the neural network parameters; (b)
profiled likelihood around the expectation value for the parameter of
interest of 10 trained inference-aware models and 10 trained
cross-entropy loss based models. The latter are constructed by building
a uniformly binned Poisson count likelihood of the conditional signal
probability output. All results correspond to Benchmark 2.}

\label{fig:subfigs_training}

\end{figure}

In Fig.~\ref{fig:training_dynamics}, the dynamics of inference-aware
optimisation are shown by the validation loss, which corresponds to the
approximate expected variance of parameter \(s\), as a function of the
training step for 10 random-initialised instances of the
\textsc{INFERNO} model corresponding to Benchmark 2. All inference-aware
models were trained during 200 epochs with SGD using mini-batches of
2000 observations and a learning rate \(\gamma=10^{-6}\). All the model
initialisations converge to summary statistics that provide low variance
for the estimator of \(s\) when the nuisance parameters are accounted
for.

To compare with alternative approaches and verify the validity of the
results, the profiled likelihoods obtained for each model are shown in
Fig.~\ref{fig:profile_likelihood}. The expected uncertainty if the
trained models are used for subsequent inference on the value of \(s\)
can be estimated from the profile width when
\(\Delta \mathcal{L} = 0.5\). Hence, the average width for the profile
likelihood using inference-aware training, \(16.97\pm0.11\), can be
compared with the corresponding one obtained by uniformly binning the
output of classification-based models in 10 bins, \(24.01\pm0.36\). The
models based on cross-entropy loss were trained during 200 epochs using
a mini-batch size of 64 and a fixed learning rate of \(\gamma=0.001\).

A more complete study of the improvement provided by the different
INFERNO training procedures is provided in \autoref{tab:results_table},
where the median and 1-sigma percentiles on the expected uncertainty on
\(s\) are provided for 100 random-initialised instances of each model.
In addition, results for 100 random-initialised cross-entropy trained
models and the optimal classifier and likelihood-based inference are
also included for comparison. The confidence intervals obtained using
INFERNO-based summary statistics are considerably narrower than those
using classification and tend to be much closer to those expected when
using the true model likelihood for inference. Much smaller fluctuations
between initialisations are also observed for the INFERNO-based cases.
The improvement over classification increases when more nuisance
parameters are considered. The results also seem to suggest the
inclusion of additional information about the inference problem in the
INFERNO technique leads to comparable or better results than its
omission.

\begin{table}
  \caption{Expected uncertainty on the parameter of interest $s$
    for each of the inference benchmarks considered using a cross-entropy
    trained neural network model, INFERNO customised for each problem
    and the optimal classifier and likelihood based results. The results
    for INFERNO matching each problem are shown with bold characters.}
  \label{tab:results_table}
  \centering
  \small
  \begin{tabular}{llllll}
\toprule
{} &                       Benchmark 0 &                       Benchmark 1 &                       Benchmark 2 &                       Benchmark 3 &                       Benchmark 4 \\
\midrule
NN classifier         &           $14.99^{+0.02}_{-0.00}$ &           $18.94^{+0.11}_{-0.05}$ &           $23.94^{+0.52}_{-0.17}$ &           $21.54^{+0.27}_{-0.05}$ &           $26.71^{+0.56}_{-0.11}$ \\
INFERNO 0             &  \boldmath$15.51^{+0.09}_{-0.02}$ &           $18.34^{+5.17}_{-0.51}$ &           $23.24^{+6.54}_{-1.22}$ &           $21.38^{+3.15}_{-0.69}$ &           $26.38^{+7.63}_{-1.36}$ \\
INFERNO 1             &           $15.80^{+0.14}_{-0.04}$ &  \boldmath$16.79^{+0.17}_{-0.05}$ &           $21.41^{+2.00}_{-0.53}$ &           $20.29^{+1.20}_{-0.39}$ &           $24.26^{+2.35}_{-0.71}$ \\
INFERNO 2             &           $15.71^{+0.15}_{-0.04}$ &           $16.87^{+0.19}_{-0.06}$ &  \boldmath$16.95^{+0.18}_{-0.04}$ &           $16.88^{+0.17}_{-0.03}$ &           $18.67^{+0.25}_{-0.05}$ \\
INFERNO 3             &           $15.70^{+0.21}_{-0.04}$ &           $16.91^{+0.20}_{-0.05}$ &           $16.97^{+0.21}_{-0.04}$ &  \boldmath$16.89^{+0.18}_{-0.03}$ &           $18.69^{+0.27}_{-0.04}$ \\
INFERNO 4             &           $15.71^{+0.32}_{-0.06}$ &           $16.89^{+0.30}_{-0.07}$ &           $16.95^{+0.38}_{-0.05}$ &           $16.88^{+0.40}_{-0.05}$ &  \boldmath$18.68^{+0.58}_{-0.07}$ \\
Optimal classifier    &                             14.97 &                             19.12 &                             24.93 &                             22.13 &                             27.98 \\
Analytical likelihood &                             14.71 &                             15.52 &                             15.65 &                             15.62 &                             16.89 \\
\bottomrule
\end{tabular}

\end{table}

\begin{figure}
\centering

\subfloat[different \(r\)
value]{\includegraphics[width=0.48\textwidth,height=\textheight]{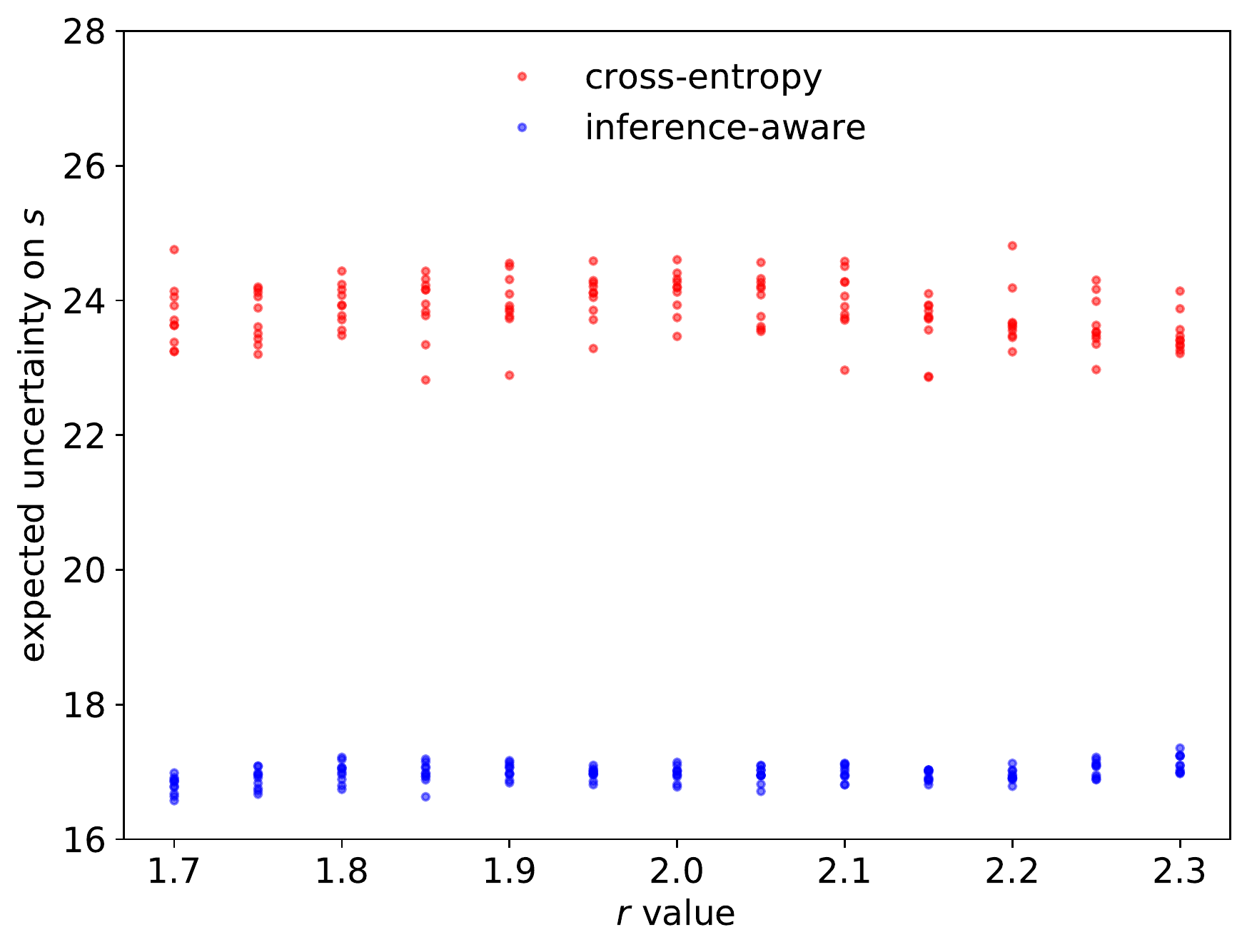}\label{fig:range_r_dist}}
\subfloat[different \(\lambda\)
value]{\includegraphics[width=0.48\textwidth,height=\textheight]{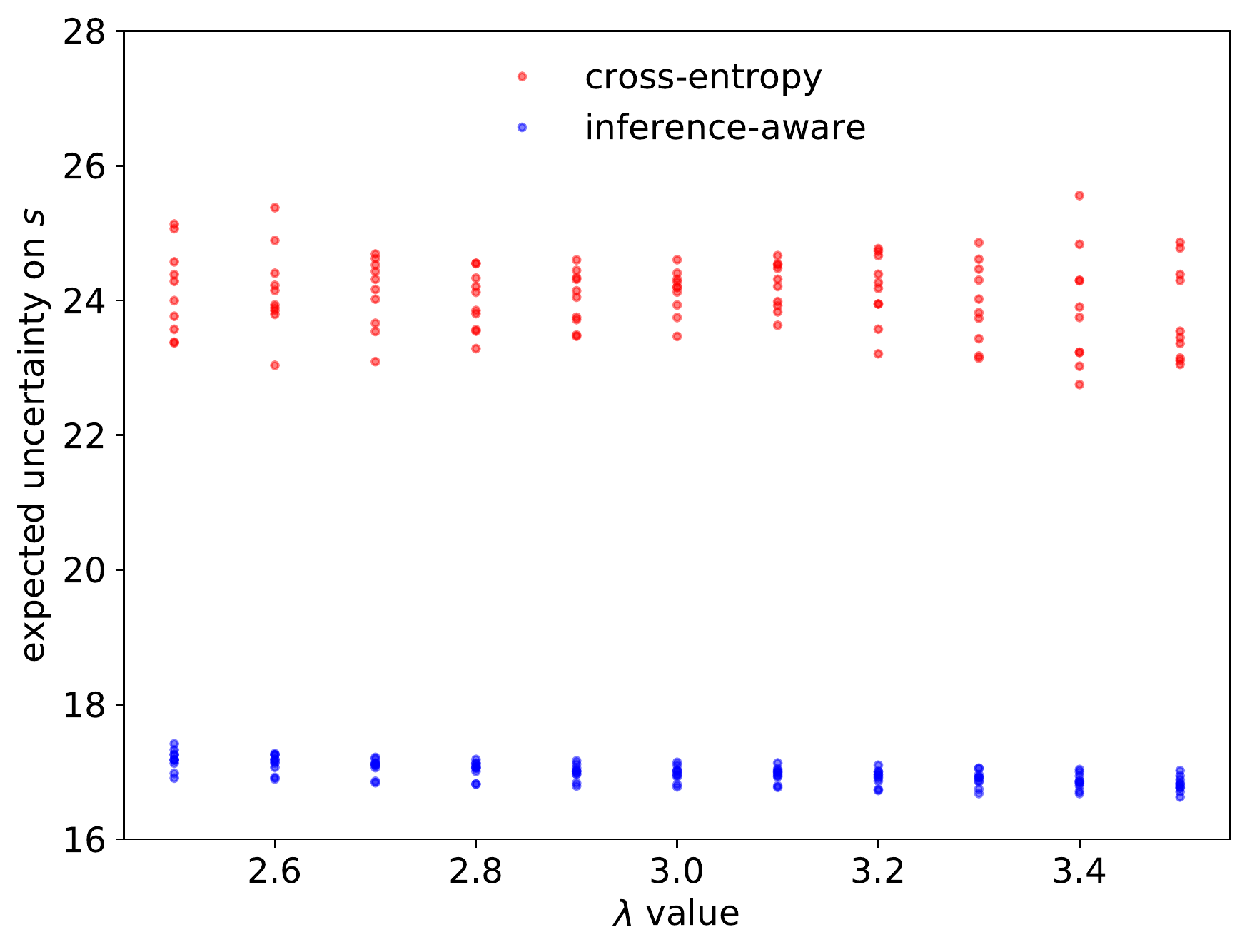}\label{fig:range_b_rate}}

\caption{Expected uncertainty when the value of the nuisance parameters
is different for 10 learnt summary statistics (different random
initialisation) based on cross-entropy classification and
inference-aware technique. Results correspond to Benchmark 2.}

\label{fig:validity_range}

\end{figure}

Given that a certain value of the parameters \(\boldsymbol{\theta}_s\)
has been used to learn the summary statistics as described in Algorithm
\autoref{alg:simple_algorithm} while their true value is unknown, the
expected uncertainty on \(s\) has also been computed for cases when the
true value of the parameters \(\boldsymbol{\theta}_{\textrm{true}}\)
differs. The variation of the expected uncertainty on \(s\) when either
\(r\) or \(\lambda\) is varied for classification and inference-aware
summary statistics is shown in Fig.~\ref{fig:validity_range} for
Benchmark 2. The inference-aware summary statistics learnt for
\(\boldsymbol{\theta}_s\) work well when
\(\boldsymbol{\theta}_{\textrm{true}} \neq \boldsymbol{\theta}_s\) in
the range of variation explored.

This synthetic example demonstrates that the direct optimisation of
inference-aware losses as those described in the Sec.~\ref{sec:method}
is effective. The summary statistics learnt accounting for the effect of
nuisance parameters compare very favourably to those obtained by using a
classification proxy to approximate the likelihood ratio. Of course,
more experiments are needed to benchmark the usefulness of this
technique for real-world inference problems as those found in High
Energy Physics analyses at the LHC.

\hypertarget{conclusions}{%
\section{Conclusions}\label{conclusions}}

Classification-based summary statistics for mixture models often suffer
from the need of specifying a fixed model of the data, thus neglecting
the effect of nuisance parameters in the learning process. The effect of
nuisance parameters is only considered downstream of the learning phase,
resulting in sub-optimal inference on the parameters of interest.

In this work we have described a new approach for building non-linear
summary statistics for likelihood-free inference that directly minimises
the expected variance of the parameters of interest, which is
considerably more effective than the use of classification surrogates
when nuisance parameters are present.

The results obtained for the synthetic experiment considered clearly
demonstrate that machine learning techniques, in particular neural
networks, can be adapted for learning summary statistics that match the
particularities of the inference problem at hand, greatly increasing the
information available for subsequent inference. The application of
INFERNO to non-synthetic examples where nuisance parameters are
relevant, such as the systematic-extended Higgs dataset
\autocite{estrade2017adversarial}, are left for future studies.

Furthermore, the technique presented can be applied to arbitrary
likelihood-free problems as long as the effect of parameters over the
simulated data can be implemented as a differentiable transformations.
As a possible extension, alternative non-parametric density estimation
techniques such as kernel density could very well be used in place
Poisson count models.

\hypertarget{acknowledgments}{%
\section*{Acknowledgments}\label{acknowledgments}}
\addcontentsline{toc}{section}{Acknowledgments}

Pablo de Castro would like to thank Daniel Whiteson, Peter Sadowski and
the other attendants of the ML for HEP meeting at UCI for the initial
feedback and support of the idea presented in this paper, as well as
Edward Goul for his interest when the project was in early stages. The
authors would also like to acknowledge Gilles Louppe and Joeri Hermans
for some useful discussions directly related to this work.

This work is part of a more general effort to develop new statistical
and machine learning techniques to use in High Energy Physics analyses
within within the AMVA4NewPhysics project, which is supported by the
European Union's Horizon 2020 research and innovation programme under
Grant Agreement number 675440. CloudVeneto is also acknowledged for the
use of computing and storage facilities provided.

\printbibliography

\appendix
\renewcommand{\thesection}{\Alph{section}}

\hypertarget{sec:sufficiency}{%
\section{Sufficient Statistics for Mixture
Models}\label{sec:sufficiency}}

Let us consider the general problem of inference for a two-component
mixture problem, which is very common in scientific disciplines such as
High Energy Physics. While their functional form will not be explicitly
specified to keep the formulation general, one of the components will be
denoted as signal \(f_s(\boldsymbol{x}| \boldsymbol{\theta})\) and the
other as background \(f_b(\boldsymbol{x} | \boldsymbol{\theta})\), where
\(\boldsymbol{\theta}\) is are of all parameters the distributions might
depend on. The probability distribution function of the mixture can then
be expressed as: \begin{equation}
p(\boldsymbol{x}| \mu, \boldsymbol{\theta} ) = (1-\mu) f_b(\boldsymbol{x} | \boldsymbol{\theta}) 
                                                + \mu f_s(\boldsymbol{x} | \boldsymbol{\theta})
\label{eq:mixture_general}\end{equation} where \(\mu\) is a parameter
corresponding to the signal mixture fraction. Dividing and multiplying
by \(f_b(\boldsymbol{x} | \boldsymbol{\theta})\) we have:
\begin{equation}
p(\boldsymbol{x}| \mu, \boldsymbol{\theta} ) = f_b(\boldsymbol{x} | \boldsymbol{\theta})   \left ( 1-\mu
                    + \mu \frac{f_s(\boldsymbol{x} | \boldsymbol{\theta})}{f_b(\boldsymbol{x} | \boldsymbol{\theta})}
                    \right )  
\label{eq:mixture_div}\end{equation} from which we can already prove
that the density ratio
\(s_{s/ b}= f_s(\boldsymbol{x} | \boldsymbol{\theta}) / f_b(\boldsymbol{x} | \boldsymbol{\theta})\)
(or alternatively its inverse) is a sufficient summary statistic for the
mixture coefficient parameter \(\mu\). This would also be the case for
the parametrization using \(s\) and \(b\) if the alternative
\(\mu=s/(s+b)\) formulation presented for the synthetic problem in Sec.
\ref{d-synthetic-mixture}.

However, previously in this work (as well as for most studies using
classifiers to construct summary statistics) we have been using the
summary statistic
\(s_{s/(s+b)}= f_s(\boldsymbol{x} | \boldsymbol{\theta}) /(  f_s(\boldsymbol{x} | \boldsymbol{\theta}) + f_b(\boldsymbol{x} | \boldsymbol{\theta}))\)
instead of \(s_{s/ b}\). The advantage of \(s_{s/(s+b)}\) is that it
represents the conditional probability of one observation
\(\boldsymbol{x}\) coming from the signal assuming a balanced mixture,
and hence is bounded between zero and one. This greatly simplifies its
visualisation and non-parametetric likelihood estimation. Taking
Eq.~\ref{eq:mixture_div} and manipulating the subexpression depending on
\(\mu\) by adding and subtracting \(2\mu\) we have: \begin{equation}
p(\boldsymbol{x}| \mu, \boldsymbol{\theta} ) = f_b(\boldsymbol{x} | \boldsymbol{\theta})   \left ( 1-3\mu
                    + \mu \frac{f_s(\boldsymbol{x} | \boldsymbol{\theta}) + f_b(\boldsymbol{x} | \boldsymbol{\theta})}{f_b(\boldsymbol{x} | \boldsymbol{\theta})}
                    \right )  
\label{eq:mixture_sub}\end{equation} which can in turn can be expressed
as:

\begin{equation}
p(\boldsymbol{x}| \mu, \boldsymbol{\theta} ) = f_b(\boldsymbol{x} | \boldsymbol{\theta})   \left ( 1-3\mu
                    + \mu \left ( 1- \frac{f_s(\boldsymbol{x} | \boldsymbol{\theta})}{f_s(\boldsymbol{x} | \boldsymbol{\theta})
                  +f_b(\boldsymbol{x} | \boldsymbol{\theta})} \right )^{-1}
                    \right )  
\label{eq:mixture_suff}\end{equation} hence proving that \(s_{s/(s+b)}\)
is also a sufficient statistic and theoretically justifying its use for
inference about \(\mu\). The advantage of both \(s_{s/(s+b)}\) and
\(s_{s/b}\) is they are one-dimensional and do not depend on the
dimensionality of \(\boldsymbol{x}\) hence allowing much more efficient
non-parametric density estimation from simulated samples. Note that we
have been only discussing sufficiency with respect to the mixture
coefficients and not the additional distribution parameters
\(\boldsymbol{\theta}\). In fact, if a subset of \(\boldsymbol{\theta}\)
parameters are also relevant for inference (e.g.~they are nuisance
parameters) then \(s_{s/(s+b)}\) and \(s_{s/b}\) are not sufficient
statistics unless the \(f_s(\boldsymbol{x}| \boldsymbol{\theta})\) and
\(f_b(\boldsymbol{x}| \boldsymbol{\theta})\) have very specific
functional form that allows a similar factorisation.

\end{document}